\documentclass[letterpaper]{article} 
\usepackage{aaai2026}  
\usepackage{times}  
\usepackage{helvet}  
\usepackage{courier}  
\usepackage[hyphens]{url}  
\usepackage{graphicx} 
\usepackage{natbib}  
\usepackage{caption} 
\usepackage{algorithm}
\usepackage{algorithmic}

\usepackage{newfloat}
\usepackage{listings}
\DeclareCaptionStyle{ruled}{labelfont=normalfont,labelsep=colon,strut=off} 
\floatstyle{ruled}
\newfloat{listing}{tb}{lst}{}
\floatname{listing}{Listing}
\usepackage{bm, comment, amsmath}
\usepackage{xcolor} 
\usepackage{multirow}

\usepackage{array} 
\usepackage{booktabs} 

\usepackage{amsfonts, amsmath, amsthm}
\usepackage{graphicx}
\usepackage{subcaption}

\usepackage{amssymb}
\usepackage{pifont}
\newcommand{\notcheckmark}{\checkmark\makebox[0pt][r]{\normalfont\symbol{'26}}}

\newtheorem{theorem}{Theorem}

\newtheorem{definition}{Definition}

\title{Boomda: Balanced Multi-objective Optimization \\ for Multimodal Domain Adaptation}
\author{
    Jun Sun\textsuperscript{\rm 1}, Xinxin Zhang\textsuperscript{\rm 2}, Simin Hong, Jian Zhu\textsuperscript{\rm 1$*$}, Xiang Gao\textsuperscript{\rm 1$*$} \\
}
\affiliations{
    \textsuperscript{\rm 1}Zhejinag Lab, Hangzhou, China\\
    \textsuperscript{\rm 2}Hangzhou Institute for Advanced Study, University of Chinese Academy of Sciences \\
    {Email: sunjun16sj@gmail.com \quad \quad \textsuperscript{$*$}Corresponding author}
}

\usepackage{bibentry}

\begin{document}

\maketitle

\begin{abstract}
Multimodal learning, while contributing to numerous success stories across various fields, faces the challenge of prohibitively expensive manual annotation. To address the scarcity of annotated data, a popular solution is unsupervised domain adaptation, which has been extensively studied in unimodal settings yet remains less explored in multimodal settings. In this paper, we investigate heterogeneous multimodal domain adaptation, where the primary challenge is the varying domain shifts of different modalities from the source to the target domain. We first introduce the information bottleneck method to learn representations for each modality independently, and then match the source and target domains in the representation space with correlation alignment. To balance the domain alignment of all modalities, we formulate the problem as a multi-objective task, aiming for a Pareto optimal solution. By exploiting the properties specific to our model, the problem can be simplified to a quadratic programming problem. Further approximation yields a closed-form solution, leading to an efficient modality-balanced multimodal domain adaptation algorithm. The proposed method features \textbf{B}alanced multi-\textbf{o}bjective \textbf{o}ptimization for \textbf{m}ultimodal \textbf{d}omain \textbf{a}daptation, termed \textbf{Boomda}. Extensive empirical results showcase the effectiveness of the proposed approach and demonstrate that Boomda outperforms the competing schemes. The code is is available at: \url{https://github.com/sunjunaimer/Boomda.git}.

\end{abstract}

%


\section{Introduction}
Multimodal learning typically leverages heterogeneous and complementary signals, such as acoustic, visual, and linguistic information, to perform machine learning tasks like classification, clustering and retrieval. Thanks to recent advances in hardware and model design, multimodal learning has been applied to various applications, including but not limited to action recognition \cite{woo2023towards}, affective computing \cite{sun2023layer, zhang2024amanda}, and medical analysis \cite{liu2023m3ae}. Compared with the unimodal alternative, multimodal learning achieves remarkable performance improvement. However, one of its notorious drawbacks is that collecting and annotating multimodal data is expensive and time-consuming. Consequently, the scarcity of annotated data presents a major challenge for the practical application of multimodal learning.

\begin{figure}[t]
	\includegraphics[width= 1\columnwidth]{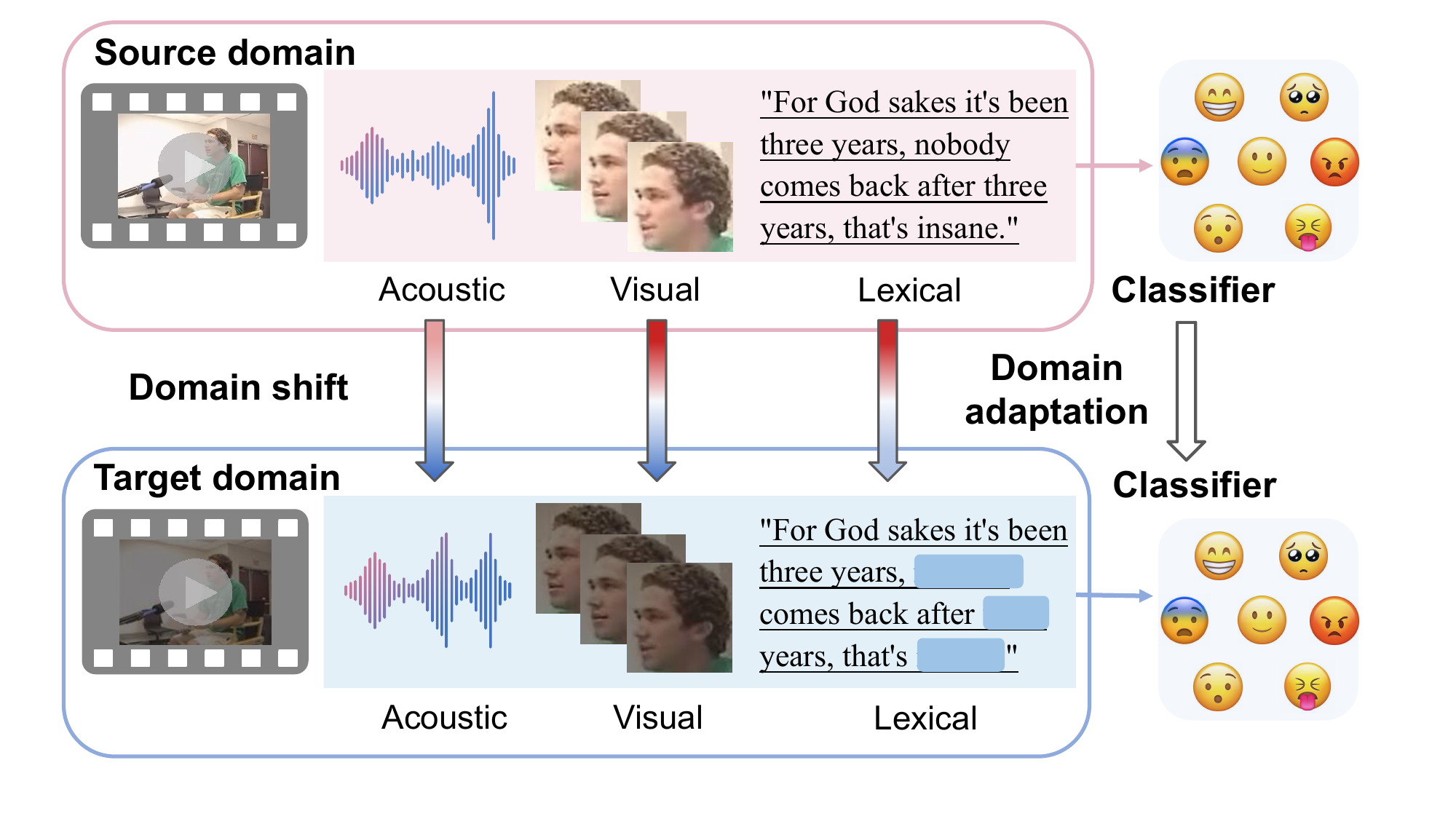} 
	\caption{Domain shift in the context of multimodal emotion recognition. The example sample is drawn from dataset IEMOCAP \cite{busso2008iemocap}.}
	\label{fig:domainshift}
	\vspace{-0.5cm}
\end{figure}

To alleviate this issue, unsupervised domain adaptation has merged as a powerful approach for enhancing the model's ability to transfer knowledge from a label-rich source domain to an unlabeled but related target domain. The objective of unsupervised domain adaptation is to train a model on the source domain while optimizing its performance on the target domain, which lacks labels during training. Mainstream approaches usually align source and target domains and learn generalizable representations via either directly minimizing the feature distribution shift between the two domains, or training an adversarial domain classifier. A massive amount of techniques following these principles have been developed, demonstrating impressive results across various tasks, including image classification \cite{hu2023open, lee2023towards}, semantic segmentation \cite{wang2023dynamically}, object detection \cite{kennerley20232pcnet}, question answering \cite{dua2023adapt}, among others. 

Nevertheless, the existing domain adaptation literature primarily focuses on unimodal settings, particularly within the areas of computer vision and natural language processing. In comparison, multimodal domain adaptation remains less investigated, yet is attracting growing interests due to the popularity of multimodal learning \cite{sun2025adversarial}. From the perspective of modality diversity, multimodal domain adaptation is conceptually divided into two categories: homogeneous and heterogeneous multimodal adaptation \cite{singhal2023domain}. The former focuses on scenarios where multiple modalities share similar underlying structure or environment. For instance, in multimodal visual domain adaptation tasks, different modalities refer to 2D image and 3D point cloud \cite{xing2023cross, liu2021adversarial}, optical flow and RGB image \cite{munro2020multi},or CT and MRI image \cite{kruse2021multi}. In such cases, the gaps between source and target domains of different modalities are relatively small, suggesting that the domains of different modalities can be aligned uniformly.

The latter, heterogeneous multimodal domain adaptation, deals with scenarios where different modalities exhibit diverse forms and reside in separate spaces. A typical example is the multimodal emotion recognition task as illustrated in Figure~\ref{fig:domainshift}, where acoustic, visual and lexical modalities are utilized simultaneously to detect emotions. Each modality faces distinct domain shift factors (e.g., background noise, illumination, and talking scenarios for the three modalities, respectively), naturally leading to varying degrees of distribution shifts from the source to the target domain across different modalities. 

In this paper, we investigate heterogeneous multimodal domain adaptation, which is more challenging than its homogeneous counterpart, as analyzed above. Directly applying current unimodal domain adaptation techniques to the heterogeneous multimodal setting potentially causes an imbalanced alignment of different modalities. The imbalance can result in some modalities dominating the training process and being well-trained, while others remaining under-trained \cite{fan2023pmr, sun2024redcore}. Henceforth, it is critical to balance the training for multimodal models in order to fully take advantage of all modalities.

Towards the goal of achieving balanced multimodal domain adaptation, this work develops a model framework building upon information bottleneck (IB) theory \cite{saxe2019information, kawaguchi2023does} to align the representations of the source and target domains, and proposes an approach to balance the alignment across modalities using the multi-objective optimization scheme. Specifically, we first construct the model with pretrained backbones for each modality. In order to retain general knowledge while adapting to new tasks and domains, the pretrained backbones are partially finetuned with some layers frozen. Then, we apply IB theory to formulate the training objective, promoting modality independence. The IB-based method ensures that each modality performs label prediction, thereby encouraging each to develop its optimal representation independently. Subsequently, the representations of each modality between the source and target domains are matched using correlation alignment (Coral) \cite{sun2016return}. 

The objectives of aligning all modalities are competing with each other, which hence can be formulated as a multi-objective optimization \cite{navon2022multi, hu2024revisiting}. To balance the alignment losses among all modalities, we employ multiple gradient descent algorithm (MGDA) \cite{sener2018multi} to obtain the Pareto optimal solution. This algorithm involves solving an optimization problem at each training iteration. Our analysis reveals that for our model structure, this optimization problem enjoys a desirable property, which facilitates the formulation of an approximated problem. Fortunately, the approximated problem admits a closed-form solution, and thus results in an efficient training algorithm.

In summary, the present paper proposes a novel approach --- \textbf{B}alanced multi-\textbf{o}bjective \textbf{o}ptimization for \textbf{m}ultimodal \textbf{d}omain \textbf{a}daptation, dubbed \textbf{Boomda}. The primary contributions are as follows:
\begin{enumerate}
	\item We develop a multimodal domain adaptation framework, where each modality is encouraged to learn its optimal representation separately based on the IB theory. Then, the source and target domains are aligned in the representation space with correlation alignment. 
	\item To balance all modalities, we formulate the alignment for them as a multi-objective optimization problem, which can be solved efficiently using MGDA algorithm via exploiting a favorable property of our problem.  
	\item Extensive experiments conducted on widely used benchmark datasets demonstrate the superior performance of Boomda compared to previous schemes.
\end{enumerate}

\section{Related Works}

\subsection{Domain adaptation}

A plethora of prior studies have been devoted to domain adaptation, and survey papers \cite{wang2018deep, li2024comprehensive} offer a comprehensive review of them. In this section, we cover the most relevant literature, which can be broadly classified into two groups: adversarial-based methods and moment-based methods. Adversarial-based methods center on learning domain-invariant features by
an adversarial scheme where a domain discriminator is trained against the feature extractor. The earliest work that demonstrates the effectiveness of adversarial learning for domain adaptation is DANN \cite{ganin2016domain}, following which numerous adversarial-based methods emerge. MDAN \cite{zhao2018adversarial} investigates multiple source domain adaptation and devises two versions of optimization strategies.    Label prediction information is introduced as conditioning for domain alignment in CDAN \cite{long2018conditional} and MADA \cite{pei2018multi}. CDA \cite{yadav2023cda} integrates contrastive learning into domain adaptation to achieve class-level alignment.
DADA \cite{tang2020discriminative} combines domain and category classifiers as a shared classifier to encourage a mutually inhibitory relation between domain and category predictions.
Similarly, DALN \cite{chen2022reusing} develops a discriminator-free adversarial model via reusing the task-specific classifier as a discriminator. 
Other adversarial methods, PCL \cite{li2024probabilistic} and DADA\cite{ren2024towards}, incorporate data augmentation from the raw feature space and representation space, respectively.

Moment-based methods mitigate domain shifts by minimizing moment-based distribution discrepancy across domains. 
Maximum mean discrepancy (MMD) based methods, such as DDC \cite{tzeng2014deep} and MK-MMD \cite{long2015learning}, represent first-order moment approaches which match the mean of the representations. DDC directly minimizes the maximum mean discrepancy of the source and target domains, and MK-MMD incorporates a multi-kernel strategy to enhance DDC. Coral \cite{ sun2016return, sun2016deep} and JDDA \cite{chen2019joint} are typical second-order moment approaches, matching the covariance of the representations. Coral aligns the correlation matrix of features from two domains. Built on Coral, JDDA aims to attain class-distinct features via additionally developing instance-based and center-based discriminative learning strategies. 

	\subsection{Imbalance in multimodal learning}
	
	The imbalance issue is naturally encountered in multimodal learning and is receiving increasing research attention. In studies \cite{wang2020makes} and \cite{peng2022balanced}, it is observed that different modalities can generalize and overfit at different rates, and some strong modalities might dominate the training and suppress other weak modalities. This inspires works \cite{wang2020makes}, \cite{wu2022characterizing} and \cite{sun2021learning, peng2022balanced, sun2024redcore} to dynamically regulate the learning rates of different modalities during training according to the overfitting estimation, the gradient, and the loss advantage of each modality, respectively. Besides this line, knowledge distillation is employed in work \cite{du2021improving} to reinforce the unimodal modules from pretrained teacher models and thus prevent modality failure. Self-distillation is utilized to balance the optimization objectives of all modalities in work \cite{shi2024passion}.

	In our work, we adopt the second-order moment matching method for representation alignment, thus focusing on balancing modalities and circumventing the difficulty in balancing the competing generative and discriminative components in adversarial-based methods. We introduce a multi-objective optimization method to balance the alignment of all modalities.

\begin{figure*}[t]
	\centering
	\includegraphics[width=0.99\linewidth]{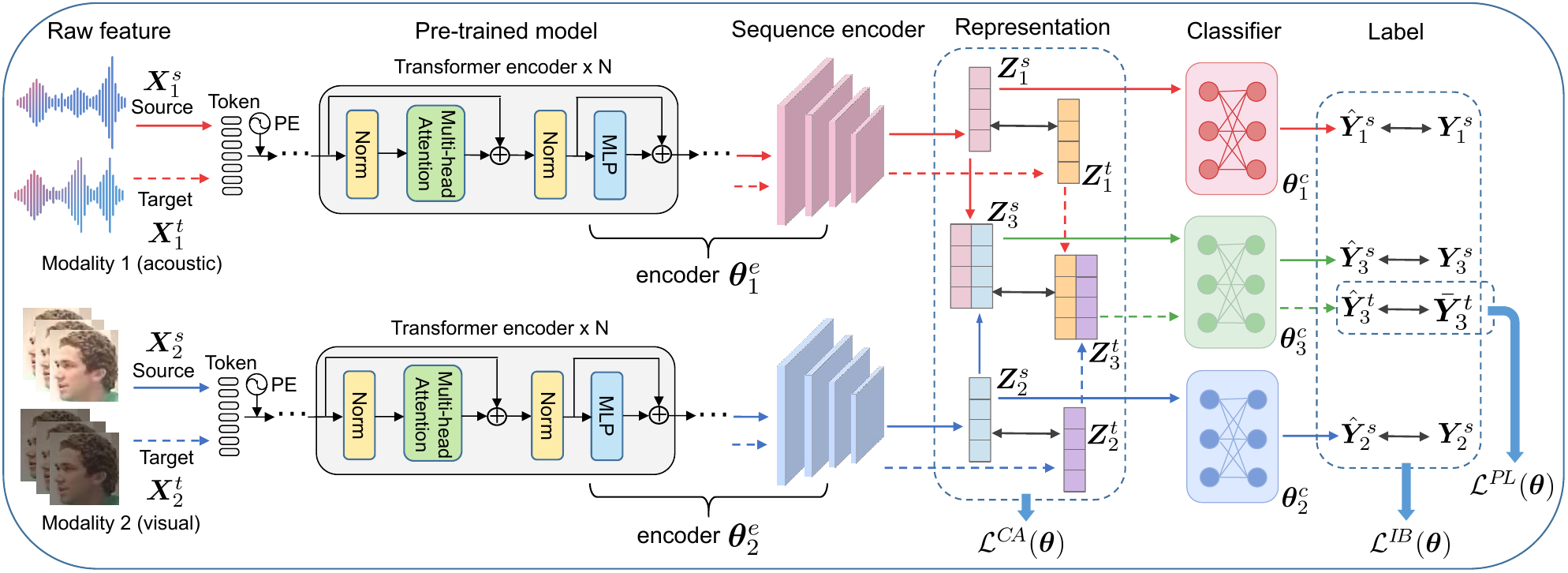} 
	\caption{Model framework with 2 modalities as an example (multimodal representation $\bm{Z}_3$ is a concatenation of $\bm{Z}_1$ and $\bm{Z}_2$; solid and dashed regular arrows represent the flows of source and target domains, respectively; double-headed arrows represent alignment or supervision signals, corresponding to the information bottleneck loss $\mathcal{L}^{I \!B}(\bm{\theta})$, pseudo label supervision loss $\mathcal{L}^{P\!L}(\bm{\theta})$ and correlation alignment loss $\mathcal{L}^{C\!A}(\bm{\theta})$).} 
	\label{fig:framework}
	\vspace{-0.5cm}
\end{figure*}

\section{Method: Boomda}

Before introducing our method, Boomda, we first define the notations to be used.

	\noindent\textbf{Notations:} Assume in a multimodal classification task, the training dataset consists of $N$ samples, each with $M$ modalities. For simplicity, we introduce an auxiliary modality containing all modalities, resulting in a total of $M+1$ modalities. Let $[I]$ for any positive integer $I$ denote the set $\{1, 2, \ldots, I\}$. The training samples are represented by $(\{\bm{x}_{n,m}\}_{m\in[M]}, \{\bm{y}_{n,m}\}_{m \in[M+1]})$, where $n\in[N]$ is the index of the samples, $\bm{x}_{n,m}\in\mathbb{R}^{d_{m}}$ represents the $d_m$-dimensional feature vector (or vector sequence) of modality $m$, $\forall m\in [M]$, and $\bm{y}_{n,m}$ denotes the label associated with modality $m$, $\forall m \in[M+1]$. In datasets where all modalities share a common label, $\bm{y}_{n,1} = \bm{y}_{n,2} = \cdots = \bm{y}_{n,M+1}$ holds.
	Suppose there are $C$ categories in the classification task; then label $\bm{y}_{n,m}$ can be either a one-hot vector or a scalar in $[C]$, and we adopt either form as necessary in the rest of the paper. For consistency, we use $\bm{x}_{n,M+1}:=[\bm{x}_{n,1}; \bm{x}_{n,2};\cdots;\bm{x}_{n,M}]$ to aggregate all modalities.
	
	Let $\bm{X}_m$ and $\bm{Y}_m$ be general feature and label random variables for all $m \in [M+1]$, with $\bm{x}_{n,m}$ and $\bm{y}_{n,m}$ being their specific instances. Let $\bm{Z}_{m}\in \mathbb{R}^d$, derived from $\bm{X}_m$, denote the representation of modality $m$; and $\bm{z}_{n,m}$ is an instance of $\bm{Z}_m$ (for brevity, we assume the representations of all modalities are $d$-dimensional vectors).  Superscripts $s$ and $t$ are used to differentiate variables from the source and target domains. For example, $\bm{X}_m^s$ and $\bm{X}_m^t$ represent the features of modality $m$ from the two domains, respectively.

In the sequel, we will first present our model design and derive the training objective function, including the model framework, IB based representation learning, pseudo labeling approach, and modality-wise representation alignment. Then, we will balance the alignment across modalities via casting it as a multi-objective optimization problem, for which we develop an efficient algorithm.

\subsection{Multimodal Domain Adaptation} 
\noindent\textbf{A. Model Framework Overview.} In this section, we focus on the model framework and ignore the implementation details, which will be elaborated later in the Numerical Results section.
Figure~\ref{fig:framework} illustrates the proposed multimodal domain adaptation framework in an example with two modalities (visual and acoustic), namely, $M=2$. The raw features $\bm{X}_m, \forall m \in[M]$ are first tokenized and fed into the pretrained transformer-based models, of which the top layers will be finetuned. Following the pretrained models are sequence encoders which further encode the sequence features into vector representations $\bm{Z}_m, \forall m\in[M]$. More formally, for each modality $m\in[M]$, the corresponding pretrained model and the sequence encoder can be summarized by a deterministic encoder function $f_m^e(\cdot;\bm{\theta}^e_m):\mathbb{R}^{d_m}\rightarrow \mathbb{R}^d$ with trainable parameter $\bm{\theta}^e_m$; then we have $\bm{Z}_M = f^e_m(\bm{X}_m;\bm{\theta}^e_m)$. The multimodal representation is denoted by $\bm{Z}_{M\!+\!1}:=[\bm{Z}_1, \bm{Z}_2, \cdots, \bm{Z}_M]$, a concatenation of the representations of all modalities.

Each modality $m\in[M+1]$ is associated with a classifier $f_m^c(\cdot, \bm{\theta}^c_m)$ with parameter $\bm{\theta}^c_m$ for label prediction; that is, $\hat{\bm{Y}}_m=f_m^c(\bm{Z}_m, \bm{\theta}^c_m)$. The multimodal prediction $\hat{\bm{Y}}_{M\! + \! 1}$ is assigned to be the ultimate predicted label. For brevity of expression, we use $\bm{\theta}:=\{\bm{\theta}^c_{M\!+\!1}\}\cup \{\bm{\theta}_m^e, \theta_m^c\}_{m\in[M]}$ to collect all model parameters, and $\bm{\theta}^e:= \{\bm{\theta}_m^e\}_{m\in[M]}$ to collect the parameters of all encoders.

\noindent\textbf{B. IB based Representation Learning.}
With the above model framework, the information follows $\bm{X}_m\rightarrow \bm{Z}_m \rightarrow \bm{Y}_m, \forall m \in [M+1]$.
Information bottleneck based representation learning aims to obtain representations $\bm{Z}^s_m, \forall m \in [M+1]$, such that they captures generalizable features, and thereby alleviates the difficulty in aligning the source and target representations.

Mathematically, the optimal representation is generated by minimizing the following IB loss on the source domain:

\begin{equation}\label{eq:IBloss}
	\mathcal{L}^{I\!B}(\bm{\theta}):=\!\!\!\!\! \sum\limits_{m\in[M+1]} \!\!\!\!\! \beta I(\bm{X}^s_m, \bm{Z}^s_m) - I(\bm{Z}^s_m, \bm{Y}_m^s),
\end{equation}
where $I(\cdot, \cdot)$ denotes the mutual information of any two random variables, and $\beta$ is a predefined coefficient. 

From the perspective of information theory, it is obvious that the resultant representation $\bm{Z}^s_m$ retains minimal information from the raw feature $\bm{X}^s_m$ yet maintains the maximal information of the label $\bm{Y}_m^s$. Therefore, $\bm{Z}^s_m$ is an optimal representation in the sense of information bottleneck theory \cite{saxe2019information, kawaguchi2023does}. Moreover, each individual modality $m$, for all $m \in [M]$, is enforced to generate its own optimal representation, which promotes modality independence and prevents some weak modalities from being dominated by strong ones.

Then, we specify how the two information terms in Eq.~\eqref{eq:IBloss} are computed.
\begin{equation}\label{eq:ixz}
	\begin{aligned}
		\!\! I(\bm{X}_m^s, \bm{Z}_m^s) \!&= H(\bm{Z}_m^s) - H(\bm{Z}_m^s|\bm{X}_m^s) \\
		& = \! H(\bm{Z}_m^s) \!=\! \mathbb{E}_{\bm{Z}_m^s}[-\log p(\bm{Z}_m^s)],
	\end{aligned}
\end{equation}
where $H(\cdot)$ represents entropy, and $H(\bm{Z}_m^s|\bm{X}_m^s)=0$ since $\bm{Z}_m = f^e_m(\bm{X}_m; \bm{\theta}_m^e)$ is a deterministic function. It is a convention to assume that $p(\bm{Z}_m^s)$ follows Gaussian distribution $\mathcal{N}(\bm{\mu}^s_m, \bm{\Sigma}^s_m)$ ($\bm{\mu}^s_m\in \mathbb{R}^d$, and  $\bm{\Sigma}^s_m\in\mathbb{R}^{d\times d}$ is a diagonal matrix). As a result, we can estimate $\bm{\mu}^s_m$ and $\bm{\Sigma}^s_m$ with the representations $\bm{z}_{n,m}^s, n\in [N^s]$, and thus the entropy of $H(\bm{Z}_m^s)$ can be obtained as:
\begin{equation}\label{eq:hz}
	H(\bm{Z}_m^s)= \frac{1}{2} \log |\bm{\Sigma}^s_m| + \frac{d}{2}(1 + \log(2\pi)), 
\end{equation}
where $|\bm{\Sigma}^s_m|$ represents the determinant of $\bm{\Sigma}^s_m$. 

Similarly, $I(\bm{Z}^s_m, \bm{Y}_m^s)$ can be written as:
\begin{equation}\label{eq:izy}
	\begin{aligned}[b]
		I(\bm{Z}^s_m, &Y_m^s) = H(\bm{Y}_m^s) -  H(\bm{Y}_m^s|\bm{Z}_m^s) \\
		& = H_{Y,m}^{s} - H(\bm{Y}_m^s|\bm{Z}_m^s) \\
		& = H_{Y,m}^{s} \! + \! \frac{1}{N^s}\sum_{n=1}^{N^s} \log p(\bm{y}_{n,m}^s|\bm{z}_{n,m}^s),
	\end{aligned}
\end{equation}
where $H(\bm{Y}_m^s) = H_{Y,m}^{s}$ is a constant independent from the model parameter $\bm{\theta}$.

Combining Eqs.~\eqref{eq:IBloss}, \eqref{eq:ixz}, \eqref{eq:hz} and \eqref{eq:izy} gives the information bottleneck loss as follows (with constant terms omitted):
\begin{equation}
	\begin{aligned}
		\mathcal{L}^{I\!B}(\bm{\theta}) \! = \!\!\!\sum_{m=1}^{M+1} \!&  \bigl[ \frac{\beta}{2} \log |\bm{\Sigma}^s_m| \\& - \frac{1}{N^s} \! \! \sum\limits_{n\in [N^s]}\!\! \log p(\bm{y}_{n,m}^s|\bm{z}_{n,m}^s) \bigr],
	\end{aligned}
\end{equation}
where the first term is a regularization for the representation that suppresses the noisy and ineffective information; and the second term corresponds to the negative log-likelihood of the prediction (equivalent to cross-entropy loss).

\noindent\textbf{C. Pseudo Labeling by Voting.} We introduce a voting strategy to yield pseudo labels for the target domain samples to supervise label prediction. Given the predictions $\hat{\bm{y}}_{n,m}^t$ of all modalities $m \in [M+1]$ (here $\hat{\bm{y}}_{n,m}^t$ is supposed to be one-hot vector), the voting result is:
\begin{equation}
	\hat{\bm{y}}^t_{n} = \sum\limits_{m\in[M\!+\!1]}{\hat{\bm{y}}_{n,m}^t}.
\end{equation}
The $c$-th element of $\hat{\bm{y}}^t_{n}$ is $(\hat{\bm{y}}^t_{n})_c, \forall c \in [C]$, denoting the number of votes for sample $n$ being classified as category $c$.

Subsequently, to ensure the reliability of labeling, only the samples that receive at least $M_v$ votes for the same category are selected to constitute the pseudo labeling set $\mathcal{N}_v^t$:
\begin{equation}\label{eq:plset}
	\mathcal{N}_v^t = \{n | \text{max} \{(\hat{\bm{y}}^t_{n})_1, \cdots, (\hat{\bm{y}}^t_{n})_C\} \geq M_v, n\in [N^t]\}.
\end{equation}
The pseudo labels for the selected samples are attained as:
\begin{equation}\label{eq:pl}
	\bar{\bm{y}}^t_{n} = \text{argmax}_c \{(\hat{\bm{y}}^t_{n})_c | c=1, 2, \cdots, C \}, \forall n \in \mathcal{N}_v^t, 
\end{equation}
where $\bar{\bm{y}}^t_{n}$ can also adopt the corresponding one-hot vector form when necessary.

Then the pseudo labels serve as supervision signals for training on the target domain, meaning the following cross-entropy loss is minimized:
\begin{equation}
			\mathcal{L}^{P\!L}(\bm{\theta}) = \frac{1}{|\mathcal{N}_v^t|}\sum\limits_{n \in \mathcal{N}_v^t}\sum\limits_{c\in [C]} - (\bar{\bm{y}}^t_{n})_c  \text{log}  (\hat{\bm{y}}^t_{n,M \! + \! 1})_c,
\end{equation}
where $|\cdot|$ represents the cardinality of a set, and $(\cdot)_c$ is the $c$-th element of a vector.

\noindent\textbf{D. Representation Alignment.} We align the source and target domains for each modality in its own representation space separately. Specifically, we first calculate the correlation matrices of $\bm{Z}_m^s$ and $\bm{Z}_m^t$, denoting them as $\bm{C}_m^s$ and $\bm{C}_m^t$, respectively, for $m \in [M+1]$. Then, we match the representations by minimizing the following correlation alignment (CA) loss \cite{ sun2016return}:
\begin{equation}
	\mathcal{L}_m^{C \! A}(\bm{\theta}) = || \bm{C}_m^t - \bm{C}_m^s ||_F^2, 
\end{equation}
where $||\cdot||_F$ means the Frobenius norm. Let  $\bm{L}^{C\!A}(\bm{\theta}):=[\mathcal{L}_1^{C\!A}(\bm{\theta}), \mathcal{L}_2^{C\!A}(\bm{\theta}), \cdots, \mathcal{L}_{M \! + \! 1}^{C\!A}(\bm{\theta})]^T$ collect the representation alignment losses for all modalities. Then, the overall loss for model training writes:
\begin{equation}\label{eq：overallloss}
	\!\! \mathcal{L}(\bm{\theta}) \!=\! \mathcal{L}^{I\!B}\!(\bm{\theta}) + \alpha_1 \mathcal{L}^{P \! L}(\bm{\theta}) + \alpha_2 h(\bm{L}^{C \! A}(\bm{\theta})),
\end{equation}
where $\alpha_1$ and $\alpha_2$ are two constant coefficients balancing the losses. Note that without incurring confusion, the formula in Eq.~\eqref{eq：overallloss} is not mathematically rigorous, since $h(\bm{L}^{C \! A}(\bm{\theta}))$ can be a vector-valued function as will be shown later.

A straightforward method to aggregate the alignment losses of all modalities is to define $h(\bm{L}^{C \! A}(\bm{\theta}))$ as a weighted sum of the losses. However, the downsides come in order: 1) without prior expert knowledge, it is difficult to specify the weights for all modalities; 2) searching for optimal weights becomes expensive when dealing with a large number of modalities; 3) predetermined fixed weights might not adapt well to the dynamic nature of alignment losses during training. To address these limitations, in the sequel, we will cast the multimodal alignment problem as a multi-objective optimization problem and develop an efficient algorithm to solve it.

\begin{table*}[]
	\centering
	\begin{tabular}{c!{\vrule width 1pt}ccccc||ccccc}
		\noalign{\hrule height 1pt} 
		\multirow{2}{*}{\textbf{Methods}} & \multicolumn{5}{c||}{\textbf{IEMOCAP}}                                                                                                                                  & \multicolumn{5}{c}{\textbf{MSP-IMPROV}}                                                                                                                            \\ \cline{2-11} 
		& \multicolumn{1}{c|}{\textbf{AL}}    & \multicolumn{1}{c|}{\textbf{AV}}    & \multicolumn{1}{c|}{\textbf{VL}}    & \multicolumn{1}{c|}{\textbf{AVL}}   & \textbf{ave.}  & \multicolumn{1}{c|}{\textbf{AL}}   & \multicolumn{1}{c|}{\textbf{AV}}    & \multicolumn{1}{c|}{\textbf{VL}}   & \multicolumn{1}{c|}{\textbf{AVL}}   & \textbf{ave.} \\ \noalign{\hrule height 1pt} 
		\textbf{D.T.}                      & \multicolumn{1}{c|}{30.99}          & \multicolumn{1}{c|}{39.34}          & \multicolumn{1}{c|}{36.37}          & \multicolumn{1}{c|}{40.31}          & 36.75          & \multicolumn{1}{c|}{25.09}         & \multicolumn{1}{c|}{34.48}          & \multicolumn{1}{c|}{35.77}         & \multicolumn{1}{c|}{38.76}          & 33.53         \\ \hline
		\textbf{DANN}                     & \multicolumn{1}{c|}{37.32}          & \multicolumn{1}{c|}{43.96}          & \multicolumn{1}{c|}{39.87}          & \multicolumn{1}{c|}{45.22}          & 41.59          & \multicolumn{1}{c|}{32.69}         & \multicolumn{1}{c|}{\textbf{42.44}} & \multicolumn{1}{c|}{37.72}         & \multicolumn{1}{c|}{37.41}          & 37.57         \\ \hline
		\textbf{DADA}                     & \multicolumn{1}{c|}{41.87}          & \multicolumn{1}{c|}{40.27}          & \multicolumn{1}{c|}{35.63}          & \multicolumn{1}{c|}{45.16}          & 40.73          & \multicolumn{1}{c|}{30.87}         & \multicolumn{1}{c|}{39.59}         & \multicolumn{1}{c|}{34.97}         & \multicolumn{1}{c|}{42.94}          & 37.09         \\ \hline
		\textbf{MADA}                     & \multicolumn{1}{c|}{38.17}          & \multicolumn{1}{c|}{35.35}          & \multicolumn{1}{c|}{42.32}          & \multicolumn{1}{c|}{49.85}          & 41.42          & \multicolumn{1}{c|}{29.86}         & \multicolumn{1}{c|}{41.32}          & \multicolumn{1}{c|}{43.38}         & \multicolumn{1}{c|}{44.13}          & 39.67         \\ \hline
		\textbf{DALN}                     & \multicolumn{1}{c|}{40.41}          & \multicolumn{1}{c|}{47.23}          & \multicolumn{1}{c|}{\textbf{42.99}} & \multicolumn{1}{c|}{45.58}          & 44.05          & \multicolumn{1}{c|}{30.36}         & \multicolumn{1}{c|}{35.68}          & \multicolumn{1}{c|}{42.23}         & \multicolumn{1}{c|}{40.29}          & 37.14         \\ \hline
		\textbf{PCL}                     & \multicolumn{1}{c|}{44.85}          & \multicolumn{1}{c|}{47.01}          & \multicolumn{1}{c|}{41.44}           & \multicolumn{1}{c|}{49.72}          & 45.76          & \multicolumn{1}{c|}{32.27}         & \multicolumn{1}{c|}{41.63}          & \multicolumn{1}{c|}{41.19}         & \multicolumn{1}{c|}{45.02}          & 40.02         \\ \hline
		\textbf{CDAN}                     & \multicolumn{1}{c|}{47.23}          & \multicolumn{1}{c|}{47.36}          & \multicolumn{1}{c|}{41.74}          & \multicolumn{1}{c|}{51.47}          & 46.95          & \multicolumn{1}{c|}{32.33}         & \multicolumn{1}{c|}{40.56}          & \multicolumn{1}{c|}{42.30}          & \multicolumn{1}{c|}{45.21}          & 40.07         \\ \hline
		\textbf{Boomda}                   & \multicolumn{1}{c|}{\textbf{49.81}} & \multicolumn{1}{c|}{\textbf{47.46}} & \multicolumn{1}{c|}{42.83}          & \multicolumn{1}{c|}{\textbf{54.82}} & \textbf{48.73} & \multicolumn{1}{c|}{\textbf{33.30}} & \multicolumn{1}{c|}{40.31}          & \multicolumn{1}{c|}{\textbf{45.10}} & \multicolumn{1}{c|}{\textbf{47.29}} & \textbf{41.50} \\ \noalign{\hrule height 1pt} 
	\end{tabular}
\caption{The performance comparisons (in terms of F1 score) of Boomda and the existing methods.} \label{tab:perfcom}
	\vspace{-0.4cm}
\end{table*}

\subsection{Pareto optimal balance across modalities}
\noindent\textbf{A. Multi-objective Optimization for Multimodal Domain Adaptation.}
The multimodal domain adaptation problem can be formulated as a multi-objective optimization problem, where multiple potentially competing objectives are optimized simultaneously. Concretely, the problem is formalized as minimizing a vector-valued loss:
\begin{equation} \label{eq:moo}
	\small
\min_{\bm{\theta}} h(\bm{L}^{C\!A}(\bm{\theta})):=  [\mathcal{L}_1^{C\!A}(\bm{\theta}), \mathcal{L}_2^{C\!A}(\bm{\theta}), \cdots, \mathcal{L}_{M \! + \! 1}^{C\!A}(\bm{\theta})]^T.
\end{equation}

The goal of multi-objective optimization centers on finding a Pareto optimal solution, as defined below.
\begin{definition}(Pareto Optimality) (a) A solution $\bm{\theta}$ dominates another solution $\bm{\theta}^{\prime}$ if $\bm{L}^{C\!A}(\bm{\theta})\neq \bm{L}^{C\!A}(\bm{\theta}^{\prime})$ and $\mathcal{L}^{C\!A}_m(\bm{\theta}) \leq \mathcal{L}^{C\!A}_m(\bm{\theta}^{\prime})$ hold for all $m\in \{1, 2, \cdots, M\}$;
	(b) A solution $\bm{\theta}^*$ is Pareto optimal if there exists no solution that dominates $\bm{\theta}^*$.
\end{definition}

As per works \cite{desideri2012multiple} and \cite{sener2018multi}, we adopt the multiple gradient descent algorithm (MGDA) to solve problem \eqref{eq:moo}. MGDA is built upon the Karush-Kuhn-Tucker (KKT) conditions, which are necessary conditions for the solution to be Pareto optimal and defines the Pareto stationary point. Specifically, any solution $\bm{\theta}$ is called a Pareto stationary point if there exists a vector $\bm{\gamma}=[\gamma_1, \gamma_2, \cdots, \gamma_{M \! + \! 1}]^T$, such that the following conditions are satisfied: a)  $\bm{\gamma} \geq \bm{0}$, b) $\bm{1}^T \cdot \bm{\gamma} = 1$, and c) $\sum_{m\in[M \! + \! 1]} \gamma_m \nabla_{\bm{\theta}}\mathcal{L}_m^{C\!A}(\bm{\theta}) = \bm{0}$, where $\bm{1}$ and $\bm{0}$ represent all-one and all-zero vector of proper size, respectively.

Then, to find Pareto stationary point involves solving the problem below:
\begin{equation*}\label{eq:MGDA}
	\begin{aligned}
\textbf{P1}:~~	\min_{\bm{\gamma}}&~~~ ||\sum_{m\in[M \! + \! 1]} \gamma_m \nabla_{\bm{\theta}}\mathcal{L}_m^{C\!A}(\bm{\theta})||_2^2  \\
	s.t. &~~~ \bm{\gamma} \geq \bm{0}, ~~\bm{1}^T \cdot \bm{\gamma} = 1.
	\end{aligned}
\end{equation*}
Suppose that $\bm{\gamma}^*$ is the optimal solution of problem \textbf{P1}, then two outcomes follow: 1) $\sum_{m\in[M \! + \! 1]} \gamma_m^* \nabla_{\bm{\theta}}\mathcal{L}^{C\!A}_m(\bm{\theta})=\bm{0}$, which means that the corresponding $\bm{\theta}$ is a Pareto stationary point, and 2) $\sum_{m\in[M \! + \! 1]} \gamma_m^* \nabla_{\bm{\theta}}\mathcal{L}^{C\!A}_m(\bm{\theta}) \neq \bm{0}$, which is a descent direction for all objectives, and thus can be applied to update the model parameter. Equivalently, objective $\sum_{m\in[M \! + \! 1]} \gamma_m^* \mathcal{L}^{C\!A}_m(\bm{\theta})$ is optimized, where $\bm{\gamma}^*$ is the coefficient that balances all modalities and leads to a Pareto stationary point.

Following the work of \cite{sener2018multi}, we have a more computationally efficient version of problem \textbf{P1}:
\begin{equation*}\label{eq:MGDAa}
	\begin{aligned}
	\textbf{P2}: ~~	\min_{\bm{\gamma}}&~~~ ||\sum_{m\in[M \! + \! 1]} \gamma_m \nabla_{\bm{Z}_{M \! + \! 1}}\mathcal{L}^{C\!A}_m(\bm{\theta})||_2^2   \\
		\text{s.t.} &~~~ \bm{\gamma} \geq \bm{0}, ~~\bm{1}^T \cdot \bm{\gamma} = 1. 
	\end{aligned}
\end{equation*}
Due to space limitations, we delegate the detailed derivation of problem \textbf{P2} to the appendix in the supplementary material. In contrast to problem \textbf{P1}, problem \textbf{P2} only requires the computation of $\nabla_{\bm{Z}_{M \! + \! 1}}\mathcal{L}^{C\!A}_m(\bm{\theta})$ instead of $\nabla_{\bm{\theta}}\mathcal{L}_m^{C\!A}(\bm{\theta})$, which reduces computational cost significantly, especially for deep neural networks.

\noindent\textbf{B. Efficient Balanced Multimodal Domain Adaptation.}
In this section, we further explore the properties of problem \textbf{P2} specific to our model, which will facilitate the design of an efficient modality-balanced algorithm. For brevity, we define $\bm{g}_m:=\nabla_{\bm{Z}_{m}}\mathcal{L}^{C\!A}_m(\bm{\theta})$ and $\bm{g}^m_{M \! + \! 1}:=\nabla_{\bm{Z}_{m}}\mathcal{L}^{C\!A}_{M \! + \! 1}(\bm{\theta}), \forall m \in [M]$.
Then, it can be verified that for the proposed model, the gradients take the following forms:
\begin{equation*}
	\bm{P}:=
	\begin{bmatrix}
		\nabla_{\bm{Z}_{M \! + \! 1}}\mathcal{L}^{C\!A}_1(\bm{\theta}) \\
		\nabla_{\bm{Z}_{M \! + \! 1}}\mathcal{L}^{C\!A}_2(\bm{\theta}) \\
		\vdots \\
		\nabla_{\bm{Z}_{M \! + \! 1}}\mathcal{L}^{C\!A}_{M \! + \! 1}(\bm{\theta})
	\end{bmatrix}
		\!=\!
	\begin{bmatrix}
		\bm{g}_1 &\bm{0} &\cdots &\bm{0} \\
		\bm{0} &\bm{g}_2 &\cdots &\bm{0} \\ 
		\vdots  &\vdots &\cdots    &\vdots \\
		\bm{g}^1_{M \! + \! 1} &\bm{g}^2_{M \! + \! 1} &\cdots &\bm{g}^M_{M \! + \! 1}
	\end{bmatrix}.
\end{equation*}

With the above notations, problem \textbf{P2} can be equivalently written in a more compact manner:
\begin{equation*}\label{eq:MGDAb}
	\begin{aligned}
	\textbf{P3}:~~	\min_{\bm{\gamma}}&~~~ \bm{\gamma}^T\bm{P}\bm{P}^T\bm{\gamma}= \bm{\gamma}^T\bm{Q}\bm{\gamma}  \\
		\text{s.t.} &~~~ \bm{\gamma} \geq \bm{0}, ~~\bm{1}^T \cdot \bm{\gamma} = 1, 
	\end{aligned}
\end{equation*}
where $\bm{Q} := \bm{P}\bm{P}^T=$
\begin{equation*}
	\small
\begin{bmatrix}
	\bm{g}_1^T \bm{g}_1 &0 &\cdots &0 &\bm{g}_1^T \bm{g}^1_{M \! + \! 1} \\
	0  &\bm{g}_2^T \bm{g}_2 &\cdots &0 & \bm{g}_2^T \bm{g}^2_{M \! + \! 1} \\ 
	\vdots  &\vdots &\cdots    &\vdots & \vdots\\
	0 &0 &\cdots  &\bm{g}_M^T \bm{g}_M & \bm{g}_M^T \bm{g}^M_{M \! + \! 1} \\
	\bm{g}_1^T \bm{g}^1_{M \! + \! 1} & \bm{g}_2^T \bm{g}^2_{M \! + \! 1} &\cdots & \bm{g}_M^T \bm{g}^M_{M \! + \! 1} & \sum\limits_{m=1}^M(\bm{g}^m_{M \! + \! 1})^T \bm{g}^m_{M \! + \! 1}
\end{bmatrix}.
\end{equation*}

In experiments, it is observed that the absolute values of the off-diagonal entries of $\bm{Q}$ are much smaller than the diagonal entries, indicating that matrix $\bm{Q}$ is positive definite. The immediate result is that problem \textbf{P3} is a standard quadratic programming problem and can be solved using an off-the-shelf solver. 

Furthermore, we can approximate $\bm{Q}$ with its diagonal matrix $\bm{\tilde{Q}}$, and solve the following problem:
\begin{equation*}\label{eq:MGDAc}
	\begin{aligned}
	\textbf{P4}:~~	\min_{\bm{\gamma}}&~~~ \bm{\gamma}^T\bm{\tilde{Q}}\bm{\gamma}  \\
		\text{s.t.} &~~~ \bm{\gamma} \geq \bm{0}, ~~\bm{1}^T \cdot \bm{\gamma} = 1.
	\end{aligned}
\end{equation*}
A desirable property of problem \textbf{P4} follows.
\begin{theorem}\label{thm:cfs}
 Problem \textbf{P4} admits a closed-form solution: 
 \begin{equation}\label{eq:cfs}
\bm{\gamma} = \frac{\bm{\tilde{Q}}^{-1}\bm{1}}{\bm{1}^T\bm{\tilde{Q}}^{-1}\bm{1}}. 
\end{equation}
\end{theorem}
We provide the proof of \textbf{Theorem~\ref{thm:cfs}} for completeness in the appendix. Upon obtaining  coefficient $\bm{\gamma}$, the overall loss in Eq.~\eqref{eq：overallloss} boils down to:
\begin{equation}\label{eq:overall2}
		\!\! \mathcal{L}(\bm{\theta}) \!=\! \mathcal{L}^{I\!B}\!(\bm{\theta}) + \alpha_1 \mathcal{L}^{P \! L}(\bm{\theta}) + \alpha_2 \sum\limits_{m\in [M\!+\!1]} \gamma_m \mathcal{L}_m^{C \! A}(\bm{\theta}).
\end{equation}

{
\setlength{\textfloatsep}{8pt}
\begin{algorithm}[t!]
	\caption{Balanced multimodal domain adaptation}\label{alg:boomda}
	\begin{algorithmic}[1]
		\STATE \textbf{Initialization:} initialize model parameter $\bm{\theta}^0$. \
		\FOR {$k=0$ to $K-1$ }
		\STATE (1) perform forward pass of the model, and generate the pseudo labels according to Eqs.~\eqref{eq:plset} and \eqref{eq:pl}; 
		\STATE (2) perform two backward passes of the model to obtain matrix $\bm{Q}$ and its diagonal approximation $\tilde{\bm{Q}}$;
		\STATE (3) solve quadratic problem \textbf{P3} or use the close-form solution Eq.~\eqref{eq:cfs} to attain the weight $\bm{\gamma}^k$;
		\STATE (4) calculate the overall loss $\mathcal{L}(\bm{\theta}^k)$ as Eq.~\eqref{eq:overall2};
		\STATE update the model parameter using an optimizer (e.g., Adam): $\bm{\theta}^{k+1} \leftarrow \text{optimizer}(\bm{\theta}^{k}, \eta)$.
		\ENDFOR
		\STATE \textbf{Return:} Model parameter $\bm{\theta}^{K}$.
	\end{algorithmic}
\end{algorithm}
}

To sum up, the proposed modality-balanced multimodal domain adaptation approach is outlined in \textbf{Algorithm \ref{alg:boomda}}. At each iteration, the pseudo labeling set and the associated labels are first generated. Then, problem \textbf{P3} is solved, or the close-form solution of the approximation problem \textbf{P4} is used to obtain weight $\bm{\gamma}$ (the reason for requiring only two backward passes for all $M+1$ modalities is explained in Appendix B). Next, the overall loss is calculated, and the model parameter is updated using an optimizer with learning rate $\eta$.

\begin{figure*}
	\begin{minipage}{\textwidth}
		\begin{subfigure}[b]{0.33\linewidth} 
			\centering
			\includegraphics[height=1.725in]{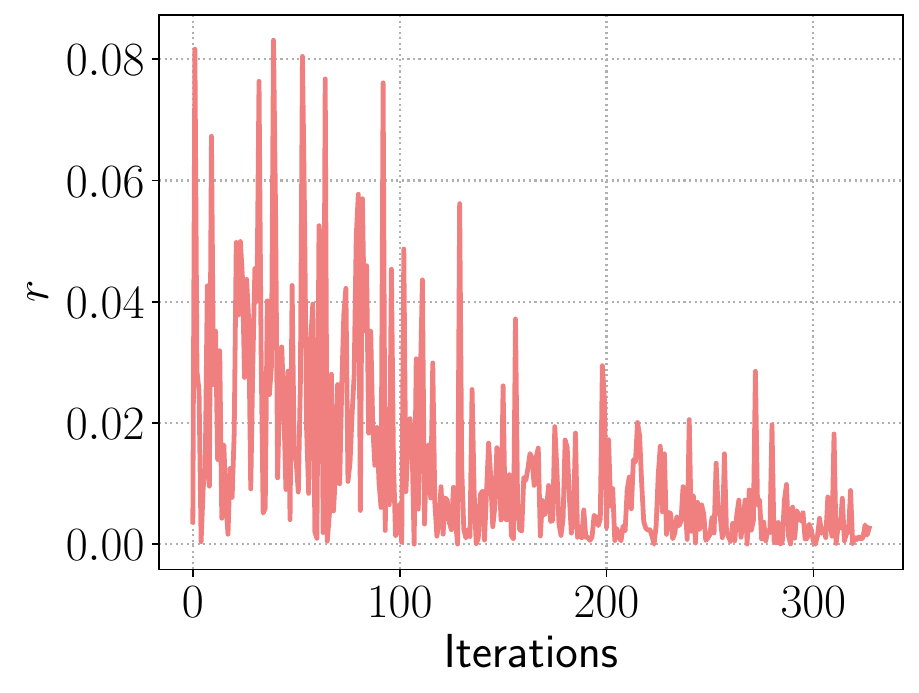}  
			\vspace{-0.1cm}
			\leftline{~~~~(a) {The characteristic (value $r$) of matrix $\bm{Q}$}.}\medskip
			\label{fig:side:a}
		\end{subfigure}%
		\begin{subfigure}[b]{0.33\linewidth} 
			\centering
			\includegraphics[height=1.725in]{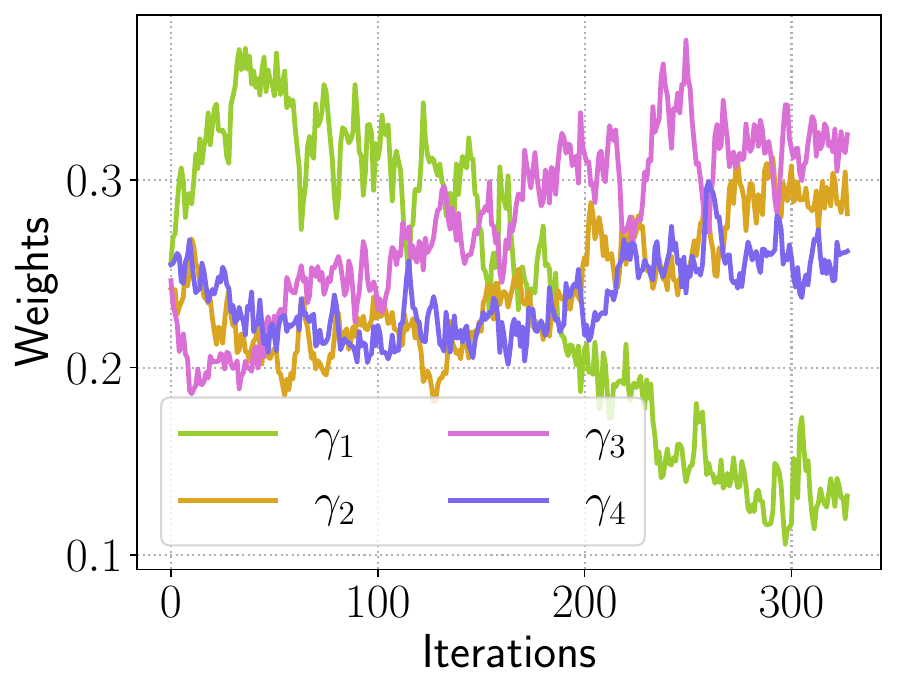}
			\vspace{-0.1cm}
			\leftline{~~~~~~~~~~~~~~~(b) {The balancing weight $\bm{\gamma}$.}}\medskip
			\label{fig:side:b}
		\end{subfigure}%
		\begin{subfigure}[b]{0.33\linewidth} 
			\centering
			\includegraphics[height=1.725in]{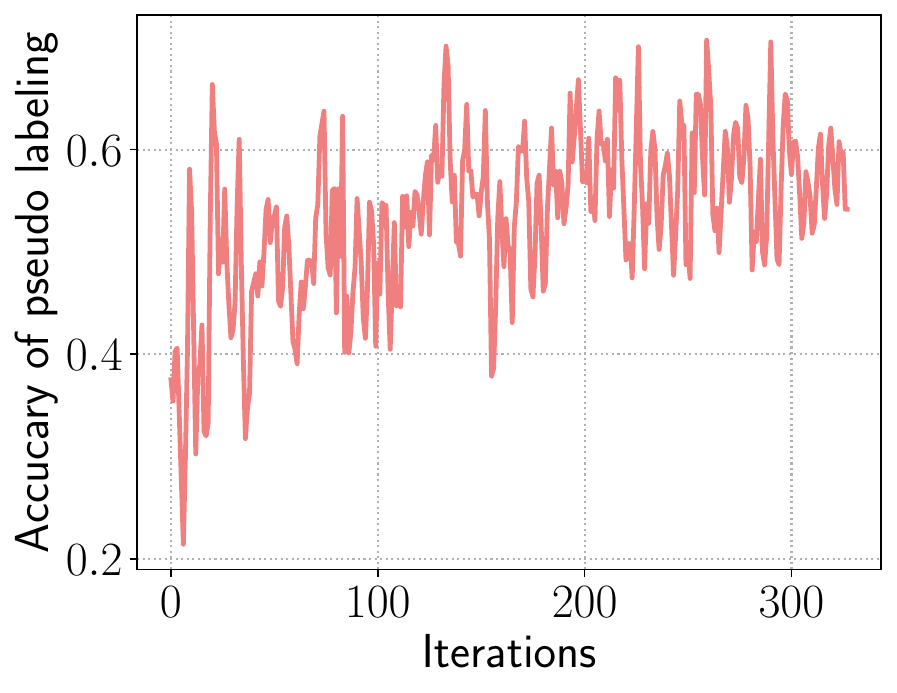}
			\vspace{-0.1cm}
			\centerline{(c) {The accuracy of pseudo labeling.}}\medskip
			\label{fig:side:c}
		\end{subfigure}%
		\vspace{-0.3cm}
		\caption{The training dynamics on the IEMOCAP dataset.}\label{fig:loss}
	\end{minipage}
	\vspace{-0.4cm}
\end{figure*}

\section{Numerical Results}
\noindent\textbf{Benchmark datasets:} We evaluate our method on the multimodal emotion recognition task with two widely used benchmark datasets, IEMOCAP \cite{busso2008iemocap} and MSP-IMPROV \cite{busso2016msp}. Both datasets contain acoustic, visual, and lexical modalities, corresponding to modalities $1$, $2$ and $3$, respectively, in our model.
IEMOCAP is composed of scripted and spontaneous dyadic conversations between actors. MSP-IMPROV features a wide range of spontaneous interactions, where participants engage in unscripted dialogues, providing rich and diverse data that reflect real-world communication scenarios.
Following work \cite{zhao2021missing}, we select samples from the four classes --- neutral, happy, sad and angry, to construct the datasets to be used in our experiments. 

For each dataset, we split it evenly and randomly into two subsets. One subset is used directly as the source domain dataset, and the other, after some manipulations, serves as the target domain dataset. Specifically, for the samples in the target domain, we inject white noise into the acoustic modality with a signal-to-noise ratio (SNR) of $1.0$; for the visual modality, the brightness of the video is reduced to 20 percent of its original level, and Gaussian noise with SNR$=0.5$ is added; for the lexical modality, 40 percent of the words in each utterance are randomly masked.

\noindent\textbf{Baseline methods:} 
In the following Comparison Studies section, we compare our model, Boomda, with DANN \cite{ganin2016domain}, CDAN \cite{long2018conditional}, MADA \cite{pei2018multi}, DALN \cite{chen2022reusing},  PCL \cite{li2024probabilistic} and DADA\cite{ren2024towards}, which are introduced in the Related Works section.

\noindent\textbf{Implementation details:}
For the acoustic modality, WavLM \cite{chen2022wavlm} followed by a TextCNN is employed as the feature encoder. For the visual modality, APViT pretrained on the RAF-DB \cite{li2017reliable} database is utilized for sequence feature extraction, and then a one-layer LSTM is utilized to encode the sequence feature. Bert-base \cite{devlin2018bert} and TextCNN are adopted for the lexical modality. The parameters in the last three layers of the pretrained models are set to be trainable, with all other parameters frozen.
The dimension of the representations $\bm{Z}_m, \forall m \in[M] $, is 256.
The Adam optimizer is used for model training with learning rate $1 \times 10^{-3}$, momentum coefficient $(0.9, 0.999)$ and batch size 48. The hyperparameter settings are $\beta=5\times 10^{-4}$, $\alpha_1= 0.5$, $\alpha_2 = 0.1$, and $M_v=3$. More implementation details can be found from the appendix and the code in the supplementary material. Model performance is measured by the weighted F1 score, averaged over three runs on four 40GB Nvidia A40 GPUs. 

\begin{table}[]
	\centering
	\begin{tabular}{cc!{\vrule width 1pt}ccccc}
		\noalign{\hrule height 1pt} 
		\textbf{CA} & \textbf{PL} & \textbf{AL}    & \textbf{AV}    & \textbf{VL}    & \textbf{AVL}   & \textbf{ave.}  \\ \noalign{\hrule height 1pt}
		\ding{55}           & \ding{55}           & 30.99          & 39.34          & 36.37          & 40.31          & 36.75          \\ \hline
		\checkmark           & \ding{55}           & 46.19          & 44.74          & 43.42          & 47.85          & 45.55          \\ \hline
		\ding{55}           & \checkmark            & 42.53          & 44.58          & \textbf{44.48} & 48.78          & 45.09          \\ \hline
		\notcheckmark         & \checkmark            & 45.96          & 45.88          & 39.91          & 53.36          & 46.27          \\ \hline
		\checkmark            & \checkmark            & \textbf{49.81} & \textbf{47.46} & 42.83          & \textbf{54.82} & \textbf{48.73} \\ \noalign{\hrule height 1pt}
	\end{tabular}
\vspace{-0.1cm}
\caption{Results of ablation studies on dataset IEMOCAP. Symbol \notcheckmark ~ in the fourth row means correlation alignment is employed without balancing the modalities; that is, coefficients $\gamma_1=\gamma_2 \cdots=\gamma_{M \!+\!1}=\frac{1}{M+1}$ are used for training.}  \label{tab:abaie}
\vspace{-0.4cm}
\end{table}

\subsection{Comparison Studies} 
The F1 score comparisons are reported in Table~\ref{tab:perfcom}, where D.T. refers to direct transfer, meaning the model is trained with only the source domain data and directly tested on the target domain samples. In the table, ``A", ``V", and ``L" represent acoustic, visual, and lexical modalities, respectively. We perform experiments with different combinations of modalities, and the ``ave." columns correspond to the average results of these combinations. On average, Boomda achieves improvements over all compared models by at least 1.78 and 1.43, on the  IEMOCAP and MSP-IMPROV datasets, respectively. More specifically, in all experiments, Boomda is only surpassed by other models in the VL setting on IEMOCAP and in the AV setting on MSP-IMPROV; except for these two settings, Boomda outperforms all other models. It is noteworthy that, without carefully balancing the modalities for domain alignment, some models can suffer from performance degradation with more modalities. For example, DALN achieves an F1 score of 47.23 with the AV (acoustic and visual) modalities on IEMOCAP, yet the F1 score decreases to 45.58 with the AVL modalities. A similar observation holds for DANN on the MSP-IMPROV dataset. With modality-balanced alignment, Boomda addresses this issue and demonstrates superior performance.
\subsection{Ablation Studies} 

In the ablation studies, we analyze the effectiveness of the two primary designs: balanced correlation alignment (CA) and pseudo labeling (PL). As presented in Table~\ref{tab:abaie}, it is clear that balanced correlation alignment and pseudo labeling each can separately improve the F1 score by over 8.00 on average (from 36.77 to 45.55 and 45.09, respectively). When these two techniques are jointly adopted, the average F1 score further increases to 48.73. Comparing the last two rows, we conclude that the balanced alignment method in Boomda promotes the F1 score by approximately 2.50.

 Moreover, we exhibit the value $r=\frac{\max\{|Q_{ij}|~|i,j \in[M+1], i\neq j\}}{\min\{Q_{ii}~ | i \in [M+1]\}}$ (the ratio of the maximum absolute value of the off-diagonal entries to the minimum diagonal entry of matrix $\bm{Q})$ in Figure~\ref{fig:loss}(a). It is demonstrated that $r$ remains small throughout the training, which justifies that we can approximate matrix $\bm{Q}$ with its diagonal matrix $\bm{\tilde{Q}}$. Figure~\ref{fig:loss}(b) illustrates how the balancing weight $\bm{\gamma}$ evolves during training: a relatively large weight is assigned to modality 1 (acoustic) at the early stage and then is reduced at the later stage. Figure~\ref{fig:loss}(c) shows the accuracy of pseudo labeling in each training batch, which presents an increasing trend with the training process. 

Due to the limited space, more experimental results, including ablation studies on MSP-IMPROV and running time comparisons, are provided in the supplementary material.

\section{Conclusions}
This paper investigates multimodal domain adaptation with a focus on balancing the alignment of different modalities. Based on the information bottleneck theory, we devise a model framework, in which each modality learns its representations and performs label prediction independently. Then, we propose a pseudo labeling strategy for target domain samples that exploits the predictions of all modalities. The target and source domains are aligned in the representation space using correlation alignment. To balance the alignment losses across modalities, we develop an efficient balanced multimodal domain adaptation approach, building upon a multi-objective optimization algorithm.


\section{Acknowledgments}
The work is supported by the National Key R\&D Program of China (Grant No. 2024YFB4505602) and the National Natural Science Foundation of China (Grant No. 62306289).

\bibliography{aaai25}

\newpage

\textbf{Supplementary Material of Paper ``Boomda: Balanced Multi-objective Optimization for Multimodal Domain Adaptation''.}
\section*{Appendix}
The numbers of figures and tables, and the references in the appendix follow those in the paper.
\subsection{A. Proof of Theorem 1}
\begin{proof}
	We first rewrite problem \textbf{P4} as follows.

	\begin{subequations}\label{eq:approx}
		\begin{align}
			\textbf{P4}: ~~\min_{\bm{\gamma}}&~~~ \bm{\gamma}^T\bm{\tilde{Q}}\bm{\gamma} \label{eq:obj} \\
			\text{s.t.} & ~~~\bm{\gamma} \geq \bm{0} \label{eq:pos} \\
			& ~~~ \bm{1}^T \cdot \bm{\gamma} = 1. \label{eq:sum}
		\end{align}
	\end{subequations}
	Let $\bm{\mu}$ and $\lambda$ denote the Lagrangian multipliers associated with constraints \eqref{eq:pos} and \eqref{eq:sum}, respectively. Then the Lagrangian function of problem \textbf{P4} writes:
	\begin{equation}
		\Gamma(\bm{\gamma}, \bm{\mu}, \lambda) = \bm{\gamma}^T\bm{\tilde{Q}}\bm{\gamma} - \bm{\mu}^T\bm{\gamma} + \lambda (\bm{1}^T\cdot \bm{\gamma}-1).
	\end{equation}
	
	Then, we have the KKT conditions of problem \textbf{P4}:
	\begin{subequations}\label{eq:approx}
		\begin{align}
			\frac{\partial \Gamma(\bm{\gamma}, \bm{\mu}, \lambda)}{\partial \bm{\gamma}} &= \bm{\tilde{Q}}\bm{\gamma} -\bm{\mu} + \lambda \bm{1} = 0; \label{eq:sta_pri}\\
			\frac{\partial \Gamma(\bm{\gamma}, \bm{\mu}, \lambda)}{\partial \lambda} &= \bm{1}^T\cdot \bm{\gamma}-1 = 0;  \label{eq:sta_dual} \\
			\bm{\gamma} &\geq 0; \label{eq:pri_ineq} \\
			\bm{\mu} &\geq 0; \label{eq:dual_ineq} \\
			\bm{\mu}^T\bm{\gamma} &= 0. \label{eq:com}
		\end{align}
	\end{subequations}
	
	We first show that $\bm{\mu} = \bm{0}$ holds. We will prove it by contradiction in the following. 
	
	Suppose that there exists an element $\mu_m$ in $\bm{\mu}$ such $\mu_m>0$. Then according to \eqref{eq:com}, the corresponding element $\gamma_m$ satisfies:
	\begin{equation}\label{eq:gamm1}
		\gamma_m = 0.
	\end{equation}
	Considering that  $\bm{\tilde{Q}}$ is a diagonal matrix, we can derive from \eqref{eq:sta_pri} that $\lambda = \mu_m > 0$. Then for any $m^{\prime} \in [M+1]$ and $m^{\prime} \neq m$, we have
	\begin{equation} \label{eq:mum}
		\mu_{m^{\prime}} = \tilde{Q}_{m^{\prime}}\gamma_{m^{\prime}} + \lambda>0, 
	\end{equation}
	where $\tilde{Q}_{m^{\prime}}$ denote the $m^{\prime}$-th diagonal entry of $\bm{\tilde{Q}}$.
	
	From Eqs. \eqref{eq:com} and \eqref{eq:mum}, one can obtain that 
	\begin{equation} \label{eq:gamm2}
		\gamma_{m^{\prime}} = 0, \forall m^{\prime} \in [M+1] ~\text{and}~ m^{\prime} \neq m.
	\end{equation}
	
	Combining Eqs. \eqref{eq:gamm1} and \eqref{eq:gamm2} leads to $\bm{\gamma} = \bm{0}$, which contradicts with \eqref{eq:sta_dual} and thus concludes that $\bm{\mu} = \bm{0}$ holds. 
	
	Upon plugging  $\bm{\mu} = \bm{0}$ into Eq.~\eqref{eq:sta_pri}, we attain:
	\begin{equation}\label{eq:ga}
		\bm{\gamma} = \lambda \bm{\tilde{Q}}^{-1} \bm{1}.
	\end{equation}
	
	Combining Eqs. \eqref{eq:ga} and \eqref{eq:sta_dual} yields the desired result:
	\begin{equation*}
		\bm{\gamma}= \frac{\bm{\tilde{Q}}^{-1}\bm{1}}{\bm{1}^T\bm{\tilde{Q}}^{-1}\bm{1}},
	\end{equation*}
	which finishes the proof.
\end{proof}

\subsection{B. Derivation of Problem \textbf{P2} from Problem \textbf{P1}}

As per works \cite{desideri2012multiple, sener2018multi}, we adopt multiple gradient descent algorithm (MGDA) to solve problem \eqref{eq:moo}. MGDA is built upon the Karush-Kuhn-Tucker (KKT) conditions, which are necessary conditions for the solution to be Pareto optimal and defines the Pareto stationary point. Specifically, any solution $\bm{\theta}$ is called a Pareto stationary point if there exists a vector $\bm{\gamma}=[\gamma_1, \gamma_2, \cdots, \gamma_{M \! + \! 1}]^T$, such that the following conditions are satisfied: a)  $\bm{\gamma} \geq \bm{0}$, b) $\bm{1}^T \cdot \bm{\gamma} = 1$, and c) $\sum_{m\in[M \! + \! 1]} \gamma_m \nabla_{\bm{\theta}}\mathcal{L}_m^{C\!A}(\bm{\theta}) = \bm{0}$, where $\bm{1}$ and $\bm{0}$ represent all-one and all-zero vector of proper size, respectively.

With these KKT conditions, an optimization problem is formulated as follows.
\begin{equation*}\label{eq:MGDA}
	\begin{aligned}
		\textbf{P1}:~~	\min_{\bm{\gamma}}&~~~ ||\sum_{m\in[M \! + \! 1]} \gamma_m \nabla_{\bm{\theta}}\mathcal{L}_m^{C\!A}(\bm{\theta})||_2^2  \\
		s.t. &~~~ \bm{\gamma} \geq \bm{0}, ~~\bm{1}^T \cdot \bm{\gamma} = 1.
	\end{aligned}
\end{equation*}
Suppose that $\bm{\gamma}^* = [\gamma_1^*, \gamma_2^*, \cdots, \gamma_{M \! + \! 1}^*]$ is the optimal solution of problem \textbf{P1}, then two outcomes follows according to the resultant $\sum_{m\in[M \! + \! 1]} \gamma_m^* \nabla_{\bm{\theta}}\mathcal{L}^{C\!A}_m(\bm{\theta})$. 1) $\sum_{m\in[M \! + \! 1]} \gamma_m^* \nabla_{\bm{\theta}}\mathcal{L}^{C\!A}_m(\bm{\theta})=\bm{0}$, which means that the corresponding $\bm{\theta}$ is a Pareto stationary point, and 2) $\sum_{m\in[M \! + \! 1]} \gamma_m^* \nabla_{\bm{\theta}}\mathcal{L}^{C\!A}_m(\bm{\theta}) \neq \bm{0}$, which is a descent direction for all objectives, and can be applied to update the model parameters; equivalently, objective $\sum_{m\in[M \! + \! 1]} \gamma_m^* \mathcal{L}^{C\!A}_m(\bm{\theta})$ is optimized.

Following work of \cite{sener2018multi}, we can have an upper bound of the objective in problem \textbf{P1}:
\begin{equation}
	\begin{aligned}
		&||\sum_{m\in[M \! + \! 1]} \gamma_m \nabla_{\bm{\theta}}\mathcal{L}^{C\!A}_m(\bm{\theta})||_2^2  \\
		= &||\sum_{m\in[M \! + \! 1]} \gamma_m \frac{\partial \bm{Z}_{M \! + \! 1}}{\partial \bm{\theta}^e}\nabla_{\bm{Z}_{M \! + \! 1}}\mathcal{L}^{C\!A}_m(\bm{\theta})||_2^2  \\
		\leq & ||\frac{\partial \bm{Z}_{M \! + \! 1}}{\partial \bm{\theta}^e}||_F^2 \cdot ||\sum_{m\in[M \! + \! 1]} \gamma_m \nabla_{\bm{Z}_{M \! + \! 1}}\mathcal{L}^{C\!A}_m(\bm{\theta})||_2^2.
	\end{aligned}
\end{equation}

We can replace the objective of problem \textbf{P1} using its upper bound with term $||\frac{\partial \bm{Z}_{M \! + \! 1}}{\partial \bm{\theta}^e}||_F^2$ dropped, since $||\frac{\partial \bm{Z}_{M \! + \! 1}}{\partial \bm{\theta}^e}||_F^2$ is independent from $\bm{\gamma}$. $\nabla_{\bm{Z}_{M \! + \! 1}}\mathcal{L}^{C\!A}_m(\bm{\theta}), \forall m \in [M \! + \! 1]$ can be calculated with just two backward passes, one for  $\nabla_{\bm{Z}_{M \! + \! 1}}\mathcal{L}^{C\!A}_{M \! + \! 1}(\bm{\theta})$, and the other for $\nabla_{\bm{Z}_{M \! + \! 1}}\mathcal{L}^{C\!A}_m(\bm{\theta}), \forall m \in [M]$. This is because that $\nabla_{\bm{Z}_{m^{\prime}}}\mathcal{L}^{C\!A}_m(\bm{\theta}) = \bm{0}, \forall m, m^{\prime} \in [M], \text{and} ~m\neq m^{\prime}$, implying that the gradient  $\nabla_{\bm{Z}_{M \! + \! 1}}\mathcal{L}^{C\!A}_m(\bm{\theta}), \forall m \in [M]$ can be obtained via computing the gradient of $\sum_{m \in [M]} \mathcal{L}_m^{C\!A}(\bm{\theta})$ with respect to $\bm{Z}_{M \! + \! 1}$. 

With the above analysis, we can have an efficient alternative of problem \textbf{P1}:
\begin{equation*}\label{eq:MGDAa}
	\begin{aligned}
		\textbf{P2}: ~~	\min_{\bm{\gamma}}&~~~ ||\sum_{m\in[M \! + \! 1]} \gamma_m \nabla_{\bm{Z}_{M \! + \! 1}}\mathcal{L}^{C\!A}_m(\bm{\theta})||_2^2   \\
		\text{s.t.} &~~~ \bm{\gamma} \geq \bm{0}, ~~\bm{1}^T \cdot \bm{\gamma} = 1. 
	\end{aligned}
\end{equation*}

\begin{figure*}
	\begin{minipage}{\textwidth}
		\begin{subfigure}[b]{0.25\linewidth} 
			\centering
			\includegraphics[height=1.3in]{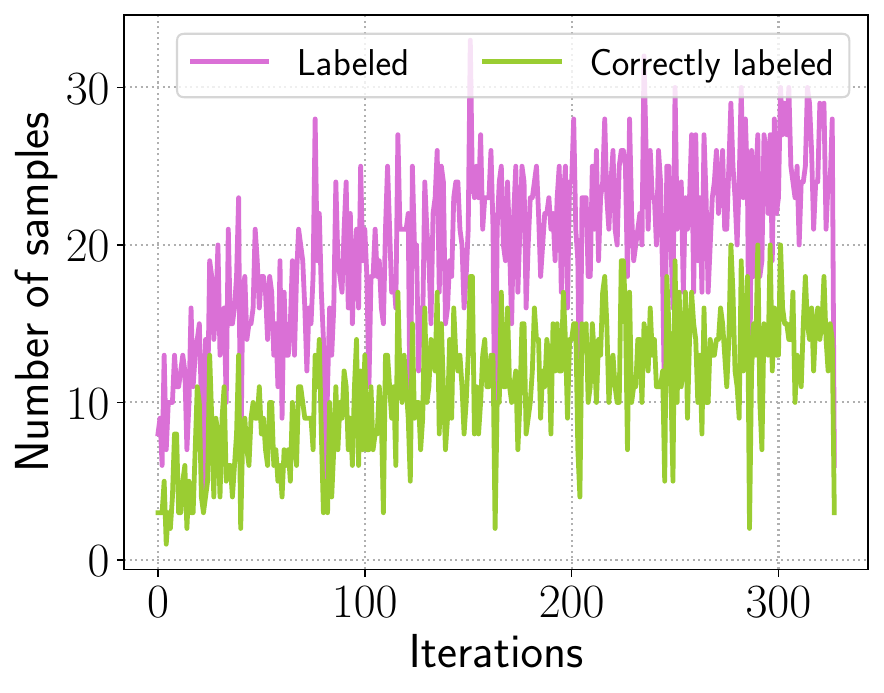}  
			\vspace{-0.1cm}
			\centerline{(a) {Number of labeled samples.}}\medskip
			\label{fig:side:a}
		\end{subfigure}%
		\begin{subfigure}[b]{0.25\linewidth} 
			\centering
			\includegraphics[height=1.3in]{figures/plr/plr_ma3b.pdf}
			\vspace{-0.1cm}
			\centerline{(b) {Accuracy of labeling.}}\medskip
			\label{fig:side:b}
		\end{subfigure}%
		\begin{subfigure}[b]{0.25\linewidth} 
			\centering
			\includegraphics[height=1.3in]{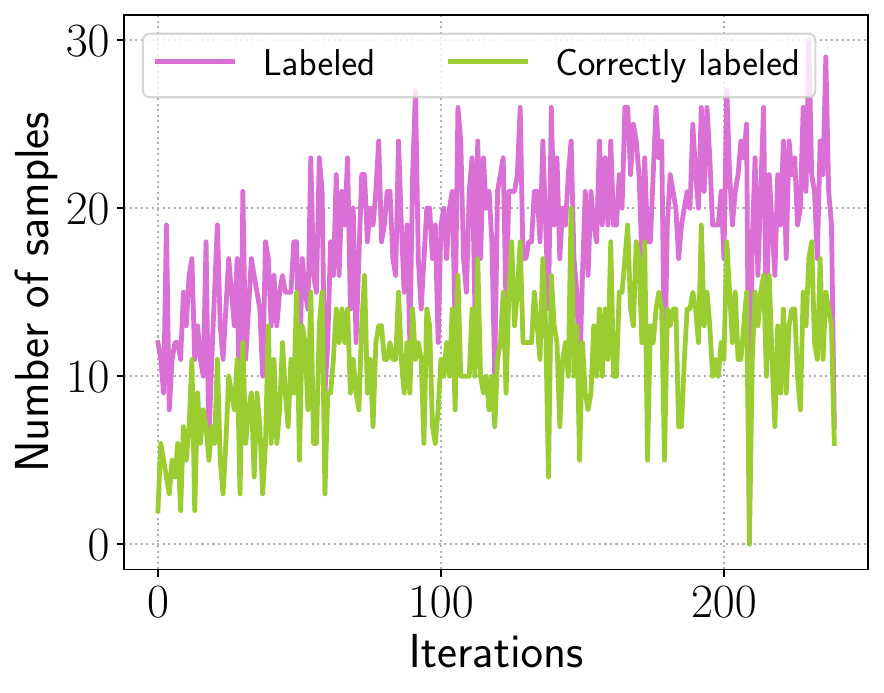}  
			\vspace{-0.1cm}
			\centerline{(c) {Number of labeled samples.}}\medskip
			\label{fig:side:c}
		\end{subfigure}%
		\begin{subfigure}[b]{0.25\linewidth} 
			\centering
			\includegraphics[height=1.3in]{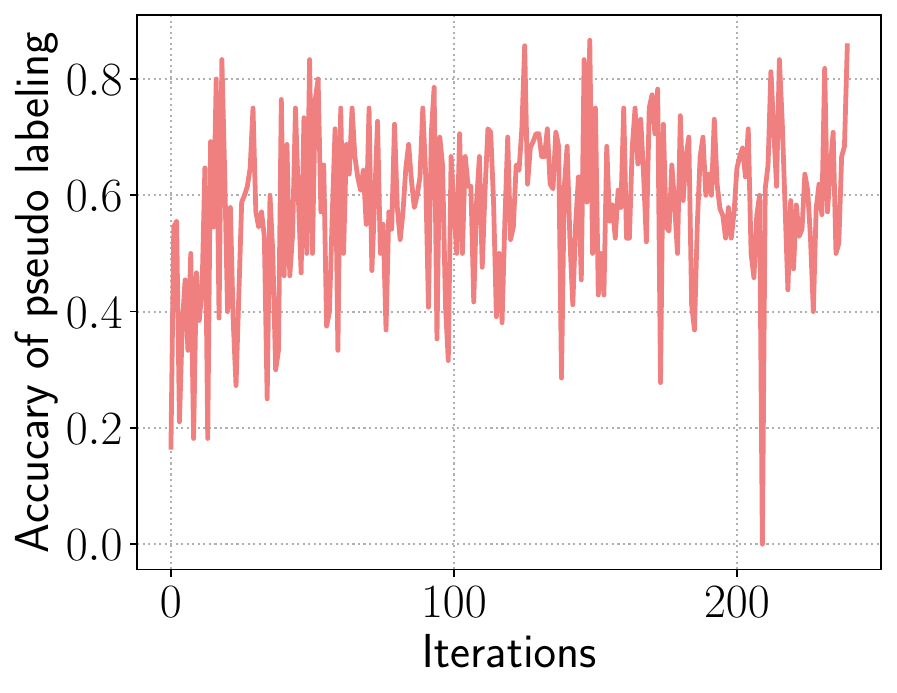}
			\vspace{-0.1cm}
			\centerline{(d) {Accuracy of labeling.}}\medskip
			\label{fig:side:d}
		\end{subfigure}%
		\vspace{-0.3cm}
		\caption{The pseudo labeling during training ((a) and (b) are the results on the IEMOCAP dataset; (c) and (d) are the results on the MSP-IMPROV dataset).}\label{fig:pl}
	\end{minipage}
	\vspace{-0.4cm}
\end{figure*}

\subsection{C. Implementation Details}
\textbf{Preprocess of threee modalities}: The audio is sampled at frequency 16kHz with a receptive field of 25ms and a frame shift of 20ms. The video is first processed with MTCNN \cite{zhang2016joint} to obtain aligned faces and each frame is cropped to size of 112 × 112. For each video utterance, we evenly sample 8 frames as the raw feature of the visual modality. For the text of the target domain, all the masked words are replaced with word "MASK".

\subsection{D. Ablation Studies}

\begin{table}[]
	\centering
	\begin{tabular}{cc!{\vrule width 1pt}ccccc}
		\noalign{\hrule height 1pt}
		\textbf{CA} & \textbf{PL} & \textbf{AL}    & \textbf{AV}    & \textbf{VL}    & \textbf{AVL}   & \textbf{ave.}  \\ \noalign{\hrule height 1pt}
		\ding{55}           & \ding{55}  & 25.09          & 34.48          & 35.77          & 38.76          & 33.53          \\ \hline
		\checkmark           & \ding{55}  & 26.73          & 40.80 & 41.69          & 42.43          & 37.91          \\ \hline
		\ding{55}           & \checkmark  & 31.12          & 39.91          & 42.61          & 47.21          & 40.21          \\ \hline
		\notcheckmark         & \checkmark  & 30.79          & \textbf{42.69}          & 42.97          & 44.68          & 40.28          \\ \hline
		\checkmark            & \checkmark  & \textbf{33.30} & 40.31          & \textbf{45.10} & \textbf{47.29} & \textbf{41.50} \\ \noalign{\hrule height 1pt}
	\end{tabular}
	\caption{Results of ablation studies on the MSP-IMPROV dataset. Symbol \notcheckmark ~ in the fourth row means correlation alignment is employed without balancing the modalities.}  \label{tab:abams}
\end{table}

\noindent\textbf{Ablation Studies on the MSP-IMPROV Dataset}: The ablation studies on the MSP-IMPROV dataset yield results similar to those observed on IEMOCAP. Namely, balanced correlation alignment and pseudo labeling independently enhance F1 score by an average of over 4.00 and 7.00, respectively, increasing it from 33.53 to 37.91 and 40.21. When both techniques are applied simultaneously, the average F1 score is further improved to 41.50. Comparing the last two rows, it is evident that the balanced alignment method in Boomda results in an additional improvement of approximately 1.20.

\noindent\textbf{Pseudo Labeling}: Figure~\ref{fig:pl}(a) illustrates the number of labeled samples and the number of correctly labeled samples during training on the IEMOCAP dataset. As training progresses, more samples are selected for labeling, indicating increased agreement between modalities. Concurrently, the accuracy of labeling improves over iterations, as shown in Figure~\ref{fig:pl}(b). Similar trends are observed with the MSP-IMPROV dataset, as depicted in Figures~\ref{fig:pl}(c) and (d).

\begin{table}[]
	\centering
\begin{tabular}{c!{\vrule width 1pt}cc}
	\noalign{\hrule height 1pt}
		\textbf{}       & \textbf{IEMOCAP} & \textbf{MSP-IMPROV} \\ \noalign{\hrule height 1pt}
		\textbf{MGDA}   & 137s             & 118s                \\ \hline
		\textbf{Boomda} & 116s             & 94s                \\ \noalign{\hrule height 1pt}
	\end{tabular}
\caption{Comparisons of running time per epoch.} \label{tab:rt}
\end{table}

\subsection{E. Running Time Comparisons}
The original MGDA algorithm leverages Frank-Wolfe algorithm to solve problem \textbf{P3} at each iteration. In comparison, Boomda adopts the closed-form solution of problem \textbf{P4} as the balancing weight $\bm{\gamma}$. Therefore, Boomda is computationally more efficient than MGDA, which is validated by the running time per epoch reported in Table~\ref{tab:rt}. Specifically, Boomda reduces therunning time of each epoch by about 15\% and 20\% for datasetsIEMOCAP and MSP-IMPROV, respectively.

\end{document}